\documentclass[pdflatex,sn-mathphys-num]{sn-jnl}

\usepackage{graphicx}
\usepackage{multirow}
\usepackage{amsmath,amssymb,amsfonts}
\usepackage[title]{appendix}
\usepackage{xcolor}
\usepackage{textcomp}
\usepackage{booktabs}
\usepackage{rotating}
\usepackage{algorithmic}
\usepackage[linesnumbered,ruled,vlined]{algorithm2e}

\begin{document}

\title[Object-Centric Spatial Reasoning for Sequential Manipulation]{Learning Object-Centric Spatial Reasoning for Sequential Manipulation in Cluttered Environments}

\author*[1]{\fnm{Chrisantus} \sur{Eze}}\email{chrisantus.eze@okstate.edu}
\author[2]{\fnm{Ryan C.} \sur{Julian}}\equalcont{This work was done when Ryan C. Julian was at Google DeepMind.}
\author[1]{\fnm{Christopher} \sur{Crick}}

\affil*[1]{\orgdiv{Department of Computer Science}, \orgname{Oklahoma State University}, \orgaddress{\city{Stillwater}, \state{OK}, \country{USA}}}

\affil[2]{\orgname{Google DeepMind}, \orgaddress{\city{Mountain View}, \state{CA}, \country{USA}}}

\abstract{Sequential manipulation in clutter requires selecting which
  objects to remove and determining how to manipulate them. Learning
  both decisions jointly can complicate training and make failures
  difficult to diagnose. We present Unveiler, a modular framework that
  separates scene reasoning and object selection from action execution
  through a discrete object-index interface. Its Spatial Relationship
  Encoder (SRE) selects the best object to grasp, while a separate
  Action Decoder generates orientation-discretized push-grasp
  actions. The SRE is trained in two stages: imitation learning from
  heuristic demonstrations, followed by search-based policy
  improvement using simulated object removals. This training procedure
  requires no real-robot training data and allows the SRE to be paired
  with a different execution policy. We evaluate Unveiler in PyBullet
  simulation and on a physical Dofbot-Pro. In real-robot experiments,
  it retrieves two occluded targets in the prescribed order in 90\% of
  scenes, compared with 20--30\% for the heuristic, GPT-4o, and the
  imitation-only variant. In PyBullet, Unveiler achieves the highest
  task-completion rate across all six clutter-density and occlusion
  conditions, reaching 90.0\% under full occlusion with 6--9 objects,
  compared with 67.5\% for the heuristic and 60.0\% for
  ThinkGrasp. These results demonstrate the potential of modular,
  object-centric spatial reasoning for sequential manipulation in
  clutter.}


\maketitle

\section{Introduction}

\begin{figure}[t]
    \centering
    \includegraphics[width=\textwidth]{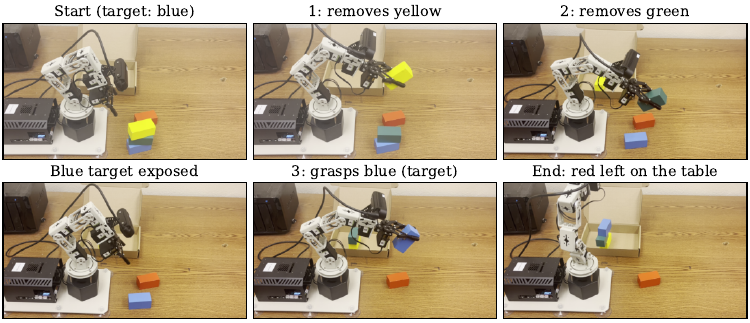}
    \caption{Unveiler on a real robot (Dofbot-Pro), retrieving a
      covered target (the blue block), seen from outside. The SRE
      selects the yellow block for removal, then the green block
      beneath it, and the robot then grasps the exposed blue
      target. The red block covers nothing and is never selected, so
      it stays on the table. The SRE reasons about object ordering
      independently of the primitive that executes each removal.}
    \label{fig:manipulation_steps}
\end{figure}

Sequential manipulation in clutter requires a robot to determine which
objects obstruct a target and in what order they should be removed for
the target to be graspable. This problem combines long-horizon spatial
reasoning with low-level action execution. Monolithic visuomotor
policies learn these decisions jointly
\cite{tang2021learning,zeng2021transporter,9815129,li2024mpgnet},
which can complicate training and make it difficult to distinguish
planning errors from execution failures. Vision-language (VLM) and
vision-language-action (VLA) models \cite{hurst2024gpt, black2024pi_0,
  bjorck2025gr00t} offer broader capabilities, but their computational
requirements and inference latency can be limiting for time-sensitive
manipulation.

We propose Unveiler, a modular framework that separates object
selection from action execution. Its Spatial Relationship Encoder
(SRE) predicts the next object to remove, while an independent Action
Decoder generates the corresponding push-grasp action. The modules
communicate through a discrete object index, allowing the selection
policy to be trained in simulation and deployed with a different
manipulation system. We train the SRE using heuristic demonstrations
followed by search-based policy improvement in simulation, without
real-robot training data. Figure~\ref{fig:manipulation_steps} shows
this modular pipeline of a scene execution in the real world.

In our real-robot evaluation, the SRE also selects objects more
reliably than GPT-4o on the tested layouts
(Table~\ref{tab:real_robot}).

Our primary contributions are:

\begin{itemize}
    \item \textbf{A compositionally decomposed architecture} that
      separates spatial reasoning from action execution via a discrete
      object-index interface. The two modules are trained
      independently, which keeps each one independently diagnosable
      and transferable: on the robot the SRE runs with a different
      arm and grasp primitive from the one used in simulation.

    \item \textbf{A transformer-based Spatial Relationship Encoder
       (SRE)} that learns to identify optimal obstacle removal sequences
       from heuristic demonstrations and from search. Trained only in
       simulation, the SRE runs
       on a real robot with no real-robot training data.

    \item \textbf{A modular two-stage training pipeline} in which
      imitation learning initializes the SRE sample-efficiently, and
      AlphaZero-style reinforcement learning then fine-tunes it:
      a search over simulated removals replaces the heuristic's
      labels, on the states the SRE itself reaches.

    \item \textbf{A real-robot evaluation} on two retrieval tasks with
      fixed, reproducible scene layouts, in which Unveiler outperforms
      the heuristic, GPT-4o, and its imitation-only variant, most
      clearly when two targets must be retrieved in order.
\end{itemize}

We train our framework entirely in simulation (PyBullet
\cite{coumans2021}, and a MuJoCo \cite{6386109} digital twin of the
robot's workspace for the real-robot experiments) and evaluate it on a
real robot and in the PyBullet simulation. Our results make a case for
the continued relevance of purpose-built compositional architectures
in an era of large-scale models: they can serve as lightweight,
interpretable, and deployable components that complement or specialize
foundation models for structured manipulation tasks.

\section{Related Work}
\label{sec:related_work}

Our work builds upon and extends existing work in robotic
manipulation, particularly in the areas of push-grasping synergy,
object-centric representation learning, and VLMs for robotics.

\subsection{Push-Grasping Synergy}
Coordinating pushing and grasping to manage clutter is well-studied
\cite{zeng2018learning, tang2021learning, deng2019deep, yang2020deep}.
TransporterNet \cite{zeng2021transporter} uses a fully convolutional
network for push-grasp visuomotor policies. More recent methods
\cite{9815129, li2024mpgnet} integrate ``move'', ``push'', and
``grasp'' actions for target-oriented grasping. Unlike these methods,
which focus on clearing clutter, Unveiler reasons about spatial
relationships between objects to determine an optimal removal sequence
via a dedicated reasoning module, rather than learning the two tasks
jointly.

\subsection{Object-centric Representations and VLMs}
Object-centric world models for manipulation
\cite{jiang2022vima,zhu2023viola,zhu2023learning,ferraro2025focus}
improve generalization and exploration by encoding per-object structure. Unveiler
specializes this toward sequential clutter: rather than targeting general-purpose manipulation, the SRE explicitly encodes long-horizon spatial
dependencies and occlusion relationships, yielding interpretable
decision-making via inspectable selection probabilities.

\subsection{Vision-Language Models for Robotic Manipulation}
The integration of large language models (LLMs) and VLMs has opened
new frontiers in robotics. Models like ThinkGrasp
\cite{qian2024thinkgrasp}, OVGNet \cite{meng2024ovgnet}, and VILG
\cite{10161041} leverage the semantic understanding of VLMs to enable
more intuitive and flexible robotic grasping.

Our work complements these advancements by proposing a lightweight and
specialized architecture. Instead of relying on a monolithic VLM for
both high-level planning and low-level control, our model uses a
focused, object-centric representation for fine-grained spatial
understanding necessary for robust low-level motor control. This
approach is not only more computationally efficient (see Table
\ref{tab:computational_analysis}) but also enhances robustness and
interpretability at the manipulation level. By bridging the gap
between high-level commands and the precise, low-level actions
required to execute them, our model offers a cheap and practical
solution for sequential manipulation in complex, real-world scenarios.

\section{Problem Formulation and Decomposition}
\label{sec:problem_statement}
We study sequential target retrieval from clutter using a
top-down RGB-D perception stack and a push-grasp primitive. Given an
RGB scene image $I$ and corresponding depth heightmap $I_g$
(constructed from the fused RGB-D point cloud), instance segmentation
produces a set of object masks $M=\{M_1,\dots,M_{N}\}$ and a target
index $t$. We use heightmaps rather than raw depth images because they
provide a robot-centric, top-down representation that reduces
perspective distortion and yields a consistent action reference frame.

\textbf{Segmentation mismatch}. In clutter, the number of detected
instances may differ from the true object count. We denote by $N$ the
number of predicted masks at the current step; Unveiler predicts and
executes one action per step, so each removal typically improves
visibility and reduces segmentation errors over time.

At each decision step, the robot must (a) choose which visible object
to remove next and (b) execute a continuous push-grasp action to
remove it. Unveiler factorizes this into a discrete object-selection
policy (SRE) and a conditional action-execution policy (Action
Decoder):
\begin{equation}
    \pi(a\mid s) \;=\; \sum_{i\in\{1,\dots,N\}} \pi_{\text{SRE}}(i\mid s, t)\;\pi_{\text{AD}}(a\mid s,i),
    \label{eq:unveiler}
\end{equation}
where the state $s$ comprises the scene observation (including $I$
and/or $I_g$) and the set of instance masks/crops $M$ with target
index $t$, the discrete choice $i$ indexes an object instance, and
$a\in\mathbb{R}^4$ parameterizes the push–grasp primitive (position,
orientation, aperture). This decomposition enables the SRE to focus on
object-centric spatial reasoning (which obstacle is most critical),
while the Action Decoder focuses on robust execution for the selected
object using a discretized orientation space.

This factorization separates object selection decisions from action
execution, allowing the two sources of failure to be examined
separately (Section~\ref{sec:theoretical_analysis}).

\section{Proposed Approach}
\label{sec:proposed_approach}

The heuristic from which our framework starts ranks the objects to
remove by two spatial criteria: (i)~distance to the workspace
boundary, which proxies grasp accessibility (peripheral objects are
less occluded and more reachable); and (ii)~distance from the target,
prioritizing objects in closest proximity to the obstructed region. By
removing peripheral objects first, scene stability is promoted
implicitly.  We decompose this into two complementary components: a
Spatial Relationship Encoder (Section~\ref{sec:encoder}) that learns
to identify the optimal next obstacle, and an Action Decoder
(Section~\ref{sec:decoder}) that executes the physical removal.

\begin{figure}
    \centering
    \includegraphics[width=1.0\linewidth]{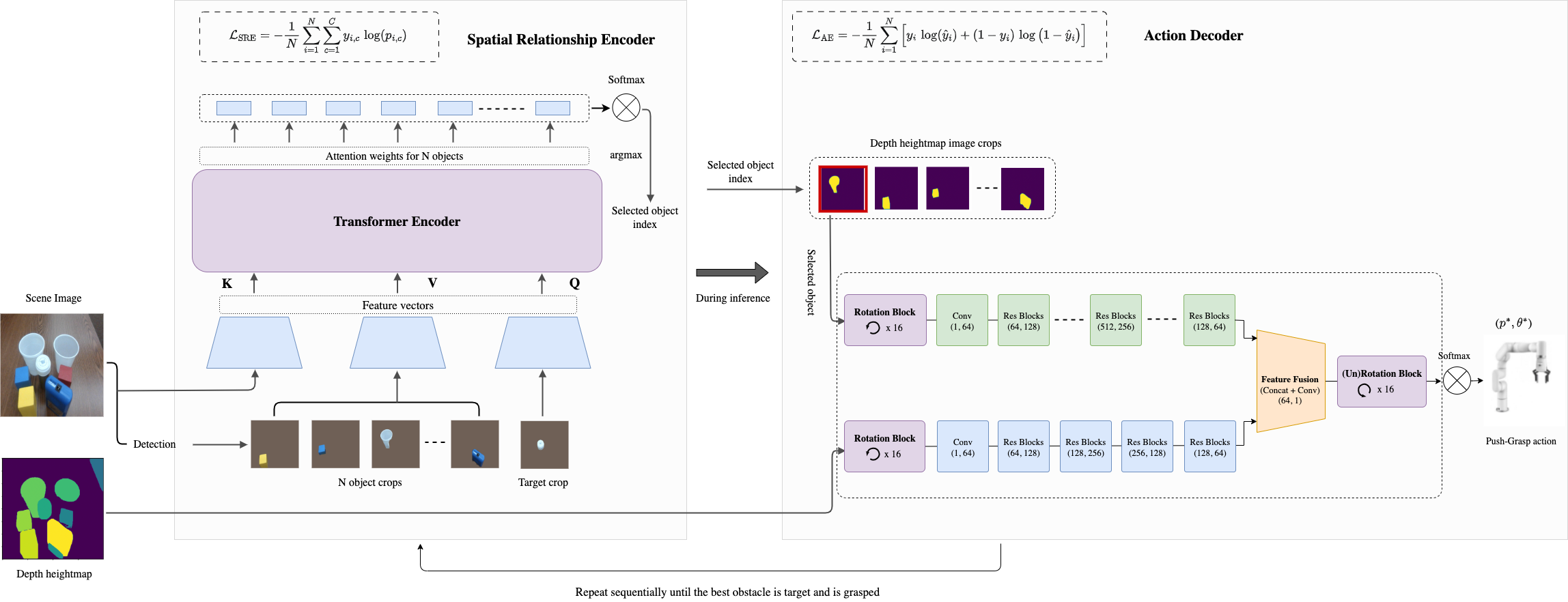}
    \caption{Unveiler system architecture. Independent training of the
      SRE and Action Decoder enables each component to specialize in
      its respective task while sharing a discrete object index at
      inference. The SRE processes scene and object-centric visual
      inputs through a transformer to select the optimal obstacle, and
      the Action Decoder generates push-grasp parameters over a
      discretized orientation space for the selected object. }
    \label{fig:arch_diagram}
\end{figure}

\subsection{Spatial Relationship Encoder}
\label{sec:encoder}
While the heuristic (Algorithm \ref{alg:heuristic_removal}) provides
effective demonstrations in moderate clutter, it operates on geometric
primitives (centroid positions and boundary distances) that become
unreliable when objects cover one another and segmentation is
imperfect. The SRE is designed to internalize the same spatial logic
into learned visual representations, gaining robustness to
segmentation noise and the ability to capture higher-order object
dependencies that the heuristic's additive scoring cannot model.

The SRE is a transformer encoder model \cite{vaswani2017attention} and
takes as input the top-camera view image of the scene, a cropped view
of the target object, and the cropped views of all visible objects in
the scene. The architecture is designed to output a discrete decision:
the index of the object that should be removed next. These indices
correspond directly to the $N$ object bounding boxes detected by the
segmentation model, where each detected object is assigned a
sequential index from 0 to $N$-1. The SRE then outputs a probability
distribution over these $N$ indices, selecting the object with the
highest removal probability. Invalid indices (those exceeding the
current object count) are automatically filtered out during inference.

The inputs are normalized and replicated across three channels to
match the expected input format. Visual features are then extracted
using a ResNet18 \cite{he2016deep} backbone pretrained on ImageNet
\cite{russakovsky2015imagenet}, transforming each \( 1 \times 144
\times 144 \) image into a \( 1 \times 1024 \) feature map for both
the target and the $N$ object crops. These features are flattened and
passed through multi-layer perceptrons (MLPs) to obtain
$1024$-dimensional latent representations. A multi-head attention
mechanism computes attention scores between the target (as query),
scene image (as key), and the surrounding objects (as value). An MLP
further processes the resulting \( N \times 512 \) attention output to
produce \( 1 \times N \) removal weights for each object.

To model inter-object dependencies and target relevance, we employ a
stack of transformer blocks. These layers enable the model to capture
higher-order relationships such as indirect occlusions (i.e., an
object indirectly blocking the target by being in front of another
obstacle).

The output of this module is a selection mask highlighting which
object should be passed to the Action Decoder for manipulation. This
discrete selection mechanism enables sequential decision making, where
only one object is removed per step, allowing the robot to reduce
occlusion and scene clutter iteratively.

\subsubsection*{\textbf{SRE Training}} The SRE is trained in two stages.
In the first stage, the policy is initialized via imitation learning
(IL) on demonstrations generated by the scoring heuristic in Algorithm
\ref{alg:heuristic_removal}. This heuristic is used to construct the
stage-1 training data. The demonstrations encode geometric removal logic
(peripheral accessibility and target proximity) into learned visual
representations, providing a sample-efficient starting point.

In the second stage, we fine-tune the IL-pretrained weights with
search-based, model-based policy improvement in the style of AlphaZero
\cite{silver2018general}. Policy improvement comes from a search over a
simulated model of the task, not from sampled returns: the search
produces improved action targets, the network is trained to reproduce
them, and the updated network then influences which states are visited
in the next round. Obstacle selection is a discrete Markov Decision
Process (MDP). The state $s_t$ is the scene, the action $a_t \in \{1,
\ldots, N\}$ is the index of the object to remove (selecting the target
means grasping it now), and every action costs one step. Because
selection is the SRE's task and execution the Action Decoder's, a
removal is simulated as ideal: the object is lifted out and the
remaining pile settles under physics.

\textbf{Search.} From any state, the simulator can be saved and
restored, so the search tries each candidate removal, lets the scene
settle, and tests whether the target has become graspable. It returns
\begin{equation}
Q(s, i) \;=\; 1 + \min_{j} Q(s_{-i}, j),
\label{eq:search}
\end{equation}
the number of steps to retrieve the target if object $i$ is removed
first, where $s_{-i}$ is the settled scene without $i$ and $Q = 1$ for
the target once it is graspable. A removal that topples or displaces
the target counts as a failure. In our real scenes, containing at most
four blocks, an episode needs at most three removals, so a search to
depth three is exact and, unlike AlphaZero, needs neither a learned
value function nor guidance from the network. Unlike the heuristic,
its labels come from simulated consequences, not from a hand-written
ranking rule.

\textbf{Training.} The SRE is trained with a cross-entropy loss towards
the soft target $\mathrm{softmax}(-Q(s,\cdot)/\tau)$, $\tau = 0.3$, on
the segmentation masks it sees at test time. The first round visits
the states along the search's own choices. In each later round the
current SRE chooses the action with probability $1-\beta$ ($\beta$
halves every round), every state it reaches is labelled by the
search, and the SRE is retrained on all states collected so far
\cite{RossGordonBagnell2011}. This exposes the SRE to its own
mistakes and to their corrections. Training runs entirely in
simulation, with no real-robot data. For the real-robot experiments,
the second stage runs in a digital twin of the robot's workspace
(Section \ref{sec:real_robot}).

\subsection{Action Decoder}
\label{sec:decoder}
Once the SRE identifies the object that should be removed, the Action
Decoder generates the low-level action required to execute the
removal. This module is responsible for determining a robust
push-grasp action, including position and rotation, tailored to the
selected object and the current scene configuration.

The Action Decoder is a fully convolutional network (FCN) that takes
two inputs: a depth heightmap of the full scene $I_g$, and a crop of
the selected object $O_i$ extracted from the depth heightmap. These
inputs are passed through separate convolutional pathways (see Figure
\ref{fig:arch_diagram}). The streams are fused to produce grasp
heatmaps across 16 discrete orientations ($22.5^\circ$ increments),
providing orientation coverage over a full rotation.

The Action Decoder is trained using a supervised cross-entropy loss
over grasp maps.  The expert used to obtain the supervision label
is shown in Algorithm \ref{alg:guided_exploration}.

\subsection{Data Collection}
The first training stage of the SRE and the Action Decoder use
demonstrations collected in the PyBullet \cite{coumans2021} simulation
environment. We develop a set of heuristic functions that identify
both the optimal obstacle to remove and its corresponding grasp
configuration at each decision step (see Algorithms
\ref{alg:heuristic_removal} and \ref{alg:guided_exploration}). We
filter demonstrations to include only successful attempts, defined by
single-object interaction and grasp stability (the object must remain
held when lifted).

\begin{algorithm}[t]
\caption{Obstacle Selection Heuristic}
\label{alg:heuristic_removal}
\begin{algorithmic}[1]
\REQUIRE Masks $M=\{M_i\}_{i=1}^{N}$, target index $t$
\ENSURE Removal order $R$
\IF{$|M| \leq 3$}
    \STATE \textbf{return} $[t]$
\ENDIF
\STATE Compute $\hat{d}_{\text{edge}}$, $\hat{d}_{\text{target}}$: normalized boundary and centroid distances
\STATE $s[i] \leftarrow \hat{d}_{\text{edge}}[i] + \hat{d}_{\text{target}}[i]$ \quad $\forall\, i \neq t$
\STATE \textbf{return} $R \leftarrow \operatorname{argsort}(s)$
\end{algorithmic}
\end{algorithm}

\begin{algorithm}[t]
\caption{Grasp Pose Heuristic}
\label{alg:guided_exploration}
\begin{algorithmic}[1]
\REQUIRE Depth heightmap $I_g$, object mask $M_i$, workspace bounds $[d_{\min}, d_{\max}]$
\ENSURE Push-grasp action $\mathbf{a} = [p_x,\, p_y,\, \theta,\, a_{\text{apt}}]^\top$
\STATE $\mathcal{V} \leftarrow \{(x,y) \mid d_{\min} \leq d(x,y) \leq d_{\max}\}$ \COMMENT{valid grasp pixels within workspace}
\IF{$\mathcal{V} = \emptyset$}
    \STATE \textbf{return} $\mathbf{0}$
\ENDIF
\STATE $\mathcal{C} \leftarrow \operatorname{contours}(M_i)$ \COMMENT{object boundary pixels}
\REPEAT
    \STATE $p_1 \sim \mathcal{U}(\mathcal{V})$ \COMMENT{sample grasp center}
    \STATE $\mathcal{P} \leftarrow \{p \in \mathcal{C} \mid \|p - p_1\|_\infty < d_{\text{push}}\}$ \COMMENT{nearby boundary points}
\UNTIL{$\mathcal{P} \neq \emptyset$}
\STATE $p_2 \sim \mathcal{U}(\mathcal{P})$ \COMMENT{push target on object boundary}
\STATE $\theta \leftarrow \operatorname{discretize}_{22.5^\circ}\!\left(-\operatorname{arctan2}(p_2 - p_1)\right)$ \COMMENT{quantized push direction}
\STATE $a_{\text{apt}} \sim \mathcal{U}(\text{aperture limits})$ \COMMENT{gripper opening width}
\STATE \textbf{return} $\mathbf{a} = [\,1.05 \cdot p_1,\ \theta,\ a_{\text{apt}}\,]^\top$
\end{algorithmic}
\end{algorithm}

\subsection{Properties of the Decomposed Policy}
\label{sec:theoretical_analysis}
The factorization equation (Eqn. \ref{eq:unveiler}) defined in Section
\ref{sec:problem_statement} is not merely an architectural
convenience. It induces two properties with direct implications for
how the system learns and how failures can be diagnosed.

\subsubsection{Object-centric inductive bias} The SRE architecture
encodes structure that a monolithic policy must discover
implicitly. The target is always the cross-attention query, anchoring
learned representations to target-relative spatial context from the
first gradient update. Per-object crops provide an object-centric
representation of each detected instance. This design is intended to
support reasoning across different object configurations;
generalization to unseen object counts and arrangements is evaluated
empirically rather than guaranteed by the crop representation alone.

\subsubsection{Separating selection and execution errors} The modular
design makes it possible to inspect object-selection errors separately
from failures during manipulation. We therefore report the fraction of
correct selections alongside episode success in the real-robot
evaluation (Table~\ref{tab:real_robot}). These metrics are related but
not interchangeable: a correct selection can still be followed by a
failed grasp, and an incorrect selection does not necessarily cause
immediate episode failure. The ablations in Table~\ref{tab:ablations}
compare architectural and training variants, but they do not establish
a general additive error bound.

\section{Experiments}
\label{sec:experiments}
We evaluated Unveiler on both real and simulation environments. On a
real robot (Section \ref{sec:real_robot}), we have two retrieval tasks
on a Dofbot-Pro, with scene layouts that are identical for every
method. We ask: (i) Does Unveiler select the right objects on a real
robot? (ii) How does it compare to the heuristic it was initialized
from and to a general-purpose VLM? (iii) What does each training stage
contribute?

\subsection{Baselines}
On the real robot, we compare against \textbf{Heur} (the heuristic of
Algorithm \ref{alg:heuristic_removal}, used to train the first stage),
\textbf{GPT-4o} \cite{hurst2024gpt}, and \textbf{IL}, the SRE after
its first training stage only. All methods share the same segmentation
and the same removal primitive, so they differ only in the object they
select. In PyBullet, we used the following state-of-the-art baselines
for object retrieval and manipulation in cluttered scenes:
\textbf{Target-Conditioned PPG} \cite{9815129}, \textbf{Heur},
\textbf{CLIP-Grounding} \cite{radford2021learning}, \textbf{GPT-4o}
\cite{hurst2024gpt}, \textbf{VILG} \cite{10161041}, and
\textbf{ThinkGrasp} \cite{qian2024thinkgrasp}.

\subsection{Real-Robot Experiments}
\label{sec:real_robot}
We evaluate on a Dofbot-Pro manipulator (Figure
\ref{fig:manipulation_steps}) equipped with a parallel gripper. The
perception system is a calibrated Intel RealSense RGB-D camera mounted
on the arm; at the home pose it views the workspace from above, at \(
640 \times 480 \) resolution.

\textbf{Training for the robot.} The second training stage runs in a
MuJoCo digital twin of the workspace: blocks of the real size and
colours, a table texture cut from real frames, and the robot's camera
view. Every rendered scene passes through the robot's own
preprocessing and segmentation, so the SRE trains on the inputs it
receives at deployment. No real-robot rollouts and no real labels are
used.

\textbf{Execution.} The module interface is a single object index, so
the SRE is deployed without the simulation's push-grasp decoder: the
selected block is removed using an off-the-shelf top-down pick-and-place
policy, which is identical for all methods. The methods therefore
differ only in which object they select.

\begin{figure}[t]
    \centering
    \includegraphics[width=0.9\linewidth]{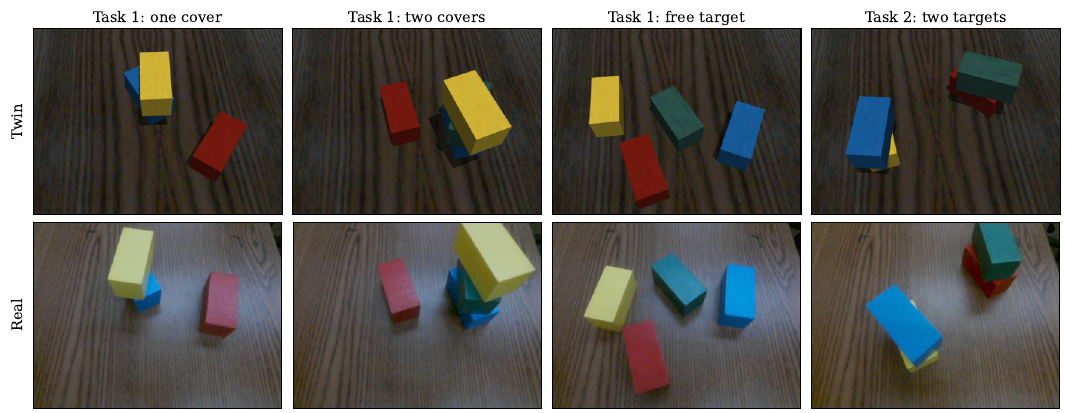}
    \caption{Evaluation scenes. Top: the layout rendered in the
      digital twin, from the arm camera. Bottom: the same layout built
      on the table. In Task~1 the target is covered by one block
      (first column), by two (second), or free (third); in Task~2
      each of two targets is covered (fourth).}
    \label{fig:real_scenes}
\end{figure}

\textbf{Tasks and matched layouts.} Scenes contain three or four
coloured blocks and are generated in the twin, then reproduced on the
table from printed layouts. Each real-robot method is evaluated on the
same set of physical layouts for a given task, rather than on a
separate set of scenes; all methods also use the same segmentation
output and the same top-down pick-and-place execution policy. Thus,
the comparison is matched at the layout and execution-policy level,
while the selected object can differ by method. In \emph{Task~1} one
target is retrieved; it is covered by one block, covered by two, or
free. In \emph{Task~2} two covered targets are retrieved in a given
order. Figure \ref{fig:real_scenes} shows one scene of each kind, in
the twin and on the table.

\textbf{Metrics.} An episode is a \textbf{success} only if every target
ends in the bin, no block is removed that was not covering a target,
and no target is grasped while still covered; a grasp at a target is
final and is not retried. This rule is stricter than retrieval alone,
because the arm can often pull a block out from under another.
\textbf{Correct selections} is the share of all selections made by a
method that either remove a block still covering a target or grasp the
current target once it is uncovered. It measures the choice,
independently of whether the grasp then succeeds. \textbf{Steps} is the
mean number of actions per episode, failed episodes included.

\begin{table}[h]
\centering
\caption{Real-robot results on the Dofbot-Pro. Sel.: correct selections (\%).}
\label{tab:real_robot}
\setlength{\tabcolsep}{4pt}
\begin{tabular}{l ccc ccc}
\toprule
\multirow{2}{*}{\textbf{Method}} &
\multicolumn{3}{c}{\textbf{Task 1}} &
\multicolumn{3}{c}{\textbf{Task 2}} \\
\cmidrule(lr){2-4}\cmidrule(lr){5-7}
 & \textbf{Success (\%)} & \textbf{Sel. (\%)} & \textbf{Steps}
 & \textbf{Success (\%)} & \textbf{Sel. (\%)} & \textbf{Steps} \\
\midrule
GPT-4o                   & 30 & 45.0 & 2.0 & 20 & 55.6 & 2.7 \\
Heur                     & 40 & 70.6 & 1.7 & 20 & 71.4 & 2.1 \\
IL                       & 40 & 69.2 & 2.6 & 30 & 75.0 & 2.8 \\
\textbf{Unveiler (Ours)} & \textbf{70} & \textbf{85.7} & 2.1
                         & \textbf{90} & \textbf{100} & 3.4 \\
\bottomrule
\end{tabular}
\end{table}

Table \ref{tab:real_robot} shows the results. On Task~2, Unveiler
retrieves both targets in 90\% of scenes, against 30\% for IL and
20\% for the heuristic and GPT-4o, and every selection it makes is
correct. On Task~1, Unveiler has the highest success rate (70\% against
at most 40\%) and the highest share of correct selections.
The baselines' lower step counts reflect episodes that end early in
failure, not more efficient retrieval.

\begin{figure}[t]
    \centering
    \includegraphics[width=\linewidth]{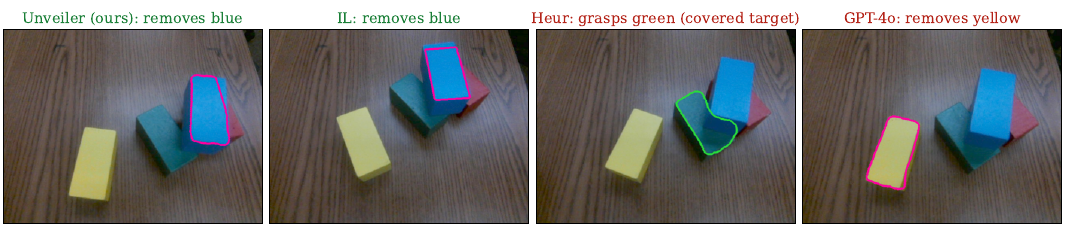}
    \caption{First selection of each method on one Task~2 scene, in
      which a blue block lies across both targets (green, then red).
      Unveiler and IL remove the blue block. The heuristic grasps the
      green target while it is still covered, and GPT-4o removes the
      yellow block, which covers nothing. IL goes on to remove the
      yellow block as well, which fails the episode.}
    \label{fig:real_first_choice}
\end{figure}

\begin{figure}[t]
    \centering
    \includegraphics[width=\linewidth]{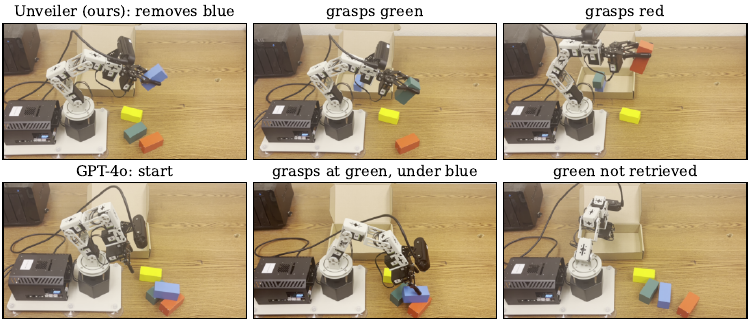}
    \caption{Top: Unveiler removes the blue block that lies across
      both targets, then grasps green and red in order. Bottom: in
      this run GPT-4o reaches for the green target while the blue
      block still covers it, and the target is not retrieved.}
    \label{fig:real_external_comparison}
\end{figure}

The failures differ by method (Figures \ref{fig:real_first_choice} and
\ref{fig:real_external_comparison}).  The heuristic mostly grasps the
target while it is still covered, and IL and GPT-4o mostly remove
blocks that are not in the way. Unveiler's four failures are two grasp
failures after correct selections, one unneeded removal, and one scene
that no method solved.

\subsection{Simulation Experiments}
\label{sec:sim_experiments}
We also evaluate the full end-to-end Unveiler framework involving both
the SRE and AD modules in the PyBullet simulation alongside other
baselines.

\begin{figure*}[h]
    \centering
    \begin{minipage}{0.42\textwidth}
        \centering
        \includegraphics[width=\linewidth]{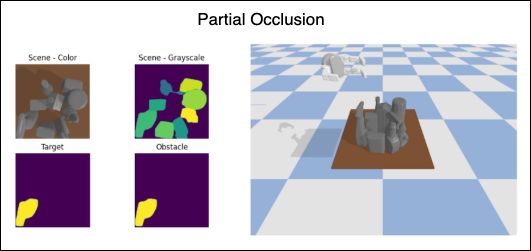}
    \end{minipage}
    \hfill
    \begin{minipage}{0.43\textwidth}
        \centering
        \includegraphics[width=\linewidth]{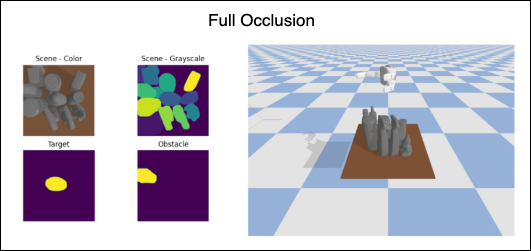}
    \end{minipage}
    \caption{Example scenes illustrating occlusion levels of the
      target object. (Left) The target object is partially occluded,
      with some visible surface area. (Right) The target object is
      fully occluded, entirely covered by surrounding objects, and not
      directly visible in the input view.}
    \label{fig:occlusion_levels}
\end{figure*}

\begin{table}[t]
\centering
\caption{Performance comparison of evaluated methods across clutter densities
  (number of objects) and occlusion levels.
  Top: Task Completion (\%, $\uparrow$). Bottom: Average Planning
  Steps ($\downarrow$). Best results per column in bold.}
\label{tab:main_sim_results}
\small
\renewcommand{\arraystretch}{1.1}
\setlength{\tabcolsep}{4.5pt}
\begin{tabular}{l cc cc cc}
\toprule
\multirow{2}{*}{\textbf{Method}} &
\multicolumn{2}{c}{\textbf{2--6}} &
\multicolumn{2}{c}{\textbf{6--9}} &
\multicolumn{2}{c}{\textbf{9--12}} \\
\cmidrule(lr){2-3}\cmidrule(lr){4-5}\cmidrule(lr){6-7}
 & \textbf{Partial} & \textbf{Full} & \textbf{Partial} & \textbf{Full} & \textbf{Partial} & \textbf{Full} \\
\midrule
\multicolumn{7}{l}{\textit{Task Completion (\%) $\uparrow$}} \\
GPT-4o          & 86.0 & 66.7 & 80.0 & 60.0 & 66.7 & 26.7 \\
CLIP-Ground.    & 80.5 & 60.3 & 67.2 & 66.7 & 53.0 & 20.0 \\
VILG            & 75.0 & 50.0 & 80.0 & 40.0 & 46.7 & 33.3 \\
ThinkGrasp      & 73.3 & 53.3 & 66.7 & 60.0 & 46.7 & 33.3 \\
PPG             & 40.6 & 50.3 & 7.5  & 4.4  & 4.4  & --   \\
Heur            & 87.3 & 67.5 & 67.5 & 47.8 & 50.0 & 20.0 \\
\textbf{Unveiler (Ours)} & \textbf{96.1} & \textbf{89.3} & \textbf{97.6} & \textbf{90.0} & \textbf{92.6} & \textbf{53.8} \\
\midrule
\multicolumn{7}{l}{\textit{Avg. Planning Steps $\downarrow$}} \\
GPT-4o          & 1.50 & 2.50 & 1.81 & 4.22 & 3.10 & \textbf{3.00} \\
CLIP-Ground.    & 1.67 & 2.11 & 3.10 & 4.20 & 4.38 & 4.67 \\
VILG            & 3.00 & 2.86 & 3.17 & 3.33 & \textbf{2.86} & 3.40 \\
ThinkGrasp      & 3.00 & 2.25 & 3.29 & 3.78 & 3.42 & 3.56 \\
PPG             & 4.42 & 4.27 & 6.00 & \textbf{3.00} & 4.00 & --   \\
Heur            & 2.35 & 2.30 & 2.55 & 3.43 & 3.40 & 3.17 \\
\textbf{Unveiler (Ours)} & \textbf{1.32} & \textbf{1.87} & \textbf{1.43} & 3.31 & \textbf{2.86} & 3.71 \\
\bottomrule
\end{tabular}
\end{table}

We validated the robustness of our system through testing on objects
of varying scales from both YCB \cite{calli2015ycb} and KIT
\cite{kasper2012kit} datasets, showing consistent performance across
different object size distributions. We categorized the objects into
two sets (seen and unseen sets) and used one for training (seen) and
the other for evaluation (unseen).

Table \ref{tab:main_sim_results} shows that Unveiler achieves the
highest task-completion rate in all six clutter-density and occlusion
conditions (see Fig. \ref{fig:occlusion_levels} for the occlusion
conditions). Its advantage is most pronounced under full occlusion
with 6–9 objects, where it achieves 90.0\% completion compared with
67.5\% for the heuristic and 60.0\% for ThinkGrasp. Performance
decreases as clutter increases, with completion reaching 53.8\% under
full occlusion with 9–12 objects. These results suggest that
sequential object selection improves retrieval performance in
cluttered scenes, although substantial failures remain in the most
challenging condition. Planning steps are considered separately
because early termination after a failed attempt can make a method
appear more efficient than it is.

\textbf{Ablation Studies}: To understand the contribution of the
individual components of Unveiler, we conducted a series of ablation
studies in the PyBullet simulation, and the results are shown in Table
\ref{tab:ablations}.

\begin{table}[ht]
\centering
\caption{Ablation study. Architectural variants evaluated at 6–9 object
  density; density generalization shown for No SRE vs.\ full Unveiler.
  Metrics: Task Completion (\%) and Average Planning Steps.}
\label{tab:ablations}
\renewcommand{\arraystretch}{1.2}
\setlength{\tabcolsep}{5pt}
\begin{tabular}{lc cccc}
\toprule
\multirow{2}{*}{\textbf{Method}} & \multirow{2}{*}{\textbf{Density}} &
\multicolumn{2}{c}{\textbf{Partial}} &
\multicolumn{2}{c}{\textbf{Full}} \\
\cmidrule(lr){3-4}\cmidrule(lr){5-6}
 & & \textbf{\%} & \textbf{Steps} & \textbf{\%} & \textbf{Steps} \\
\midrule
No SRE                      & 6--9  & 96.0          & 2.10          & 76.0          & 3.39 \\
Multi-Obstacle SRE          & 6--9  & 73.3          & 3.44          & 60.0          & 4.90 \\
Binary Mask Input           & 6--9  & 56.7          & 2.94          & 36.7          & 4.55 \\
Unveiler (Ours)             & 6--9  & \textbf{97.6} & \textbf{1.43} & \textbf{90.0} & \textbf{3.31} \\
\midrule
No SRE                      & 9--12 & 84.0          & 2.91          & 46.0          & 6.39 \\
SRE w/o Finetuning       & 9--12 & 88.4 & \textbf{2.64} & 52.0 & 4.11 \\
Unveiler (Ours)             & 9--12 & \textbf{92.6} & 2.86 & \textbf{53.8} & \textbf{3.71} \\
\bottomrule
\end{tabular}
\end{table}

\begin{itemize}
\item \textbf{SRE without second-stage fine-tuning}: At 9--12 objects,
  removing the second training stage lowers completion from 92.6\% to
  88.4\% under partial occlusion and from 53.8\% to 52.0\% under full
  occlusion. This comparison supports a benefit from the second stage
  in these conditions, although the size of the gain varies by condition.

\item \textbf{No SRE}: Replacing learned object selection with random
  selection gives mixed results at 6--9 objects: completion changes from
  97.6\% to 96.0\% under partial occlusion and from 90.0\% to 76.0\%
  under full occlusion. At 9--12 objects, the corresponding changes are
  from 92.6\% to 84.0\% and from 53.8\% to 46.0\%. These differences
  suggest that learned selection is most helpful in the more challenging
  full-occlusion settings.

\item \textbf{Multi-obstacle selection}: Predicting multiple obstacles
  in one shot lowers completion at 6--9 objects from 97.6\% to 73.3\%
  under partial occlusion and from 90.0\% to 60.0\% under full
  occlusion. This comparison supports the use of sequential rather than
  one-shot selection in these settings.

\item \textbf{Binary-mask input}: Replacing object crops with binary
  masks lowers completion at 6--9 objects from 97.6\% to 56.7\% under
  partial occlusion and from 90.0\% to 36.7\% under full occlusion.
  This result indicates that the richer visual inputs are useful in the
  tested settings.
\end{itemize}

\subsection{Interpretability of the Selection}
\begin{figure}[t]
    \centering
    \includegraphics[width=0.95\linewidth]{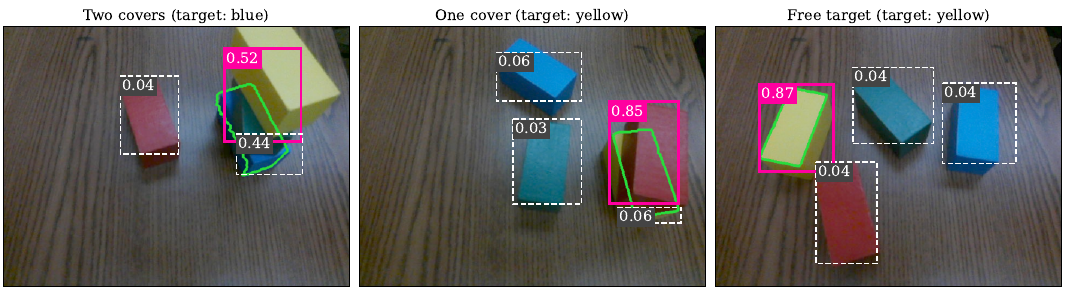}
    \caption{Selection probabilities of the SRE on real frames from
      Task~1, for every segmented object. Green: the target's outline
      from the reference frame taken before it was covered. Magenta:
      the selected object.}
    \label{fig:sre_attention_analysis}
\end{figure}

Because the SRE outputs a probability for every segmented object, its
decision can be inspected on the robot. Figure
\ref{fig:sre_attention_analysis} shows three first selections from the
Task~1 runs. With two blocks on the target, the probability is split
between the two covers (0.52 and 0.44), and the block that touches
nothing receives 0.04. With one cover, that block receives 0.85. When
the target is free, the SRE selects the target itself (0.87) and
leaves the other three blocks alone. In each case the probability mass
lies on the blocks that actually cover the target, not on the nearest
or the largest object.

\subsection{Computational Efficiency Analysis}
 Unveiler achieves strong computational efficiency as shown in Table
 \ref{tab:computational_analysis}. The reported $\sim$260\,ms for
 Unveiler is end-to-end and includes instance segmentation, crop
 extraction, SRE inference, and Action Decoder inference. The baseline
 values are estimates gathered under different inference setups, and
 the GPT-4o and ThinkGrasp measurements depend on cloud API
 latency. They should therefore be treated as indicative rather than
 as a controlled hardware-matched benchmark.

\begin{table}[h]
\centering
\caption{Computational Efficiency Comparison}
\label{tab:computational_analysis}
\begin{tabular}{lccc}
\toprule
Method & Params. (M) & Est. Inference (ms) & Relative throughput ($1\times$ = Unveiler) \\
\midrule
\textbf{Unveiler (Ours)} & 83.03 & $\sim 260$ & \textbf{1.0x} \\
VILG & 156.26 & $\sim 29,300^{a}$ & $\sim 0.0089\times$ \\
GPT-4o$^*$ & - & $> 6000^{b}$ & $<0.043\times$ \\
ThinkGrasp$^*$ & - & $> 6000^{b}$ & $<0.043\times$ \\
\botrule
\end{tabular}
\footnotetext{$^*$ Parameter count unavailable; requires cloud API inference.}
\footnotetext{$^{a}$ Average shown; actual range 2807--55,741.9 ms (approximately 0.0047--0.093$\times$ relative throughput).}
\footnotetext{$^{b}$ Dependent on API latency; range is above 6000 ms.}
\end{table}

\section{Conclusion}
We presented Unveiler, a modular framework for sequential object
retrieval in clutter that separates object selection from action
execution through a discrete object-index interface. In the real-robot
Task~2 evaluation, Unveiler retrieved two occluded targets in the
prescribed order in 90\% of scenes, compared with 20--30\% for the
heuristic, GPT-4o, and the imitation-only variant. In PyBullet
simulation, Unveiler achieved the highest task-completion rate across
all six clutter-density and occlusion conditions. Under full occlusion
with 6--9 objects, it achieved 90.0\% completion, compared with
67.5\% for the heuristic and 60.0\% for ThinkGrasp. Performance
decreased to 53.8\% under full occlusion with 9--12 objects,
indicating that dense, fully occluded scenes remain challenging.

\textbf{Implications of modularity.} The experiments support three
practical benefits of the decomposition. First, the SRE and Action
Decoder can be trained with different objectives: imitation followed by
search-based fine-tuning for selection, and supervised learning for
action generation. Second, the discrete object-index interface allows
the SRE to be deployed with a different execution policy, as demonstrated
by pairing it with a pick-and-place policy on the physical robot. Third,
selection probabilities provide a useful diagnostic signal for examining
which object the SRE prioritizes. These results motivate further study
of object-selection modules as components of larger manipulation
systems, although integration with a VLA and transfer to additional
robots remain future work.

\textbf{Limitations.} The physical evaluation uses a small set of fixed
scene layouts, and second-stage training relies on simulated object
removal, which may not capture all physical interactions. The SRE also
depends on instance segmentation, which can be unreliable when objects
overlap. On the robot, the SRE is paired with a pick-and-place policy
rather than the learned Action Decoder; therefore, the physical results
validate transfer of the selection module, not the complete learned
pipeline. Evaluating more varied layouts and integrating the Action
Decoder with the real robot are important next steps.

\backmatter

\bmhead{Supplementary information} Attached is a video of Unveiler and
GPT-4o on four of the evaluation layouts on the Dofbot-Pro, two from
each task, shown beside the arm-camera view with each selected block
outlined and, for Unveiler, its selection probability; it ends with
the real-robot results of Table~\ref{tab:real_robot}.

\section*{Declarations}

\bmhead{Data Availability Statement} The code supporting this study is openly available at \url{https://github.com/chrisantuseze/object-unveiler} and archived on Zenodo (\url{https://doi.org/10.5281/zenodo.23438694}). The release includes pretrained weights for the object segmentation model, the imitation-learning models, and the SRE after search-based policy improvement, both in simulation and in the digital twin used for the real-robot experiments. The simulated demonstration dataset is not deposited because of its size. It can be regenerated with the released data-generation script, and the search-based policy improvement can be rerun from the imitation-learning weights, as described in the repository README. The real-robot evaluation layouts and the code for the real-robot experiments are included in the repository.

\bmhead{Funding}
The authors declare that no funds, grants, or other support were received during the preparation of this manuscript.

\bibliography{unveiler}

\end{document}